\documentclass[twocolumn]{article}

\usepackage{hyperref}
\usepackage{fancyhdr}

\usepackage{graphicx}
\usepackage{tabularx}

\fancypagestyle{firstpage}{%
  \lhead{NYU Shanghai}
  \rhead{Deans' Undergraduate Research Fund}
}

\title{Exploring an LM to generate Prolog Predicates from Mathematics Questions}

\author{\href{mailto:xy2128@nyu.edu}{Xiaocheng Yang} (xy2128), \\
mentor: \href{mailto:yt2267@nyu.edu}{Yik-Cheung (Wilson) Tam} (yt2267)}

\date{\vspace{-5ex}} 

\begin{document}
\maketitle

\thispagestyle{firstpage}

\section*{Abstract}
Recently, there has been a surge in interest in NLP driven by ChatGPT. ChatGPT, a transformer-based generative language model of substantial scale, exhibits versatility in performing various tasks based on natural language. Nevertheless, large language models often exhibit poor performance in solving mathematics questions that require reasoning. Prior research has demonstrated the effectiveness of chain-of-thought prompting in enhancing reasoning capabilities. Now, we aim to investigate whether fine-tuning a model for the generation of Prolog codes, a logic language, and subsequently passing these codes to a compiler can further improve accuracy. Consequently, we employ chain-of-thought to fine-tune LLaMA7B as a baseline model and develop other fine-tuned LLaMA7B models for the generation of Prolog code, Prolog code + chain-of-thought, and chain-of-thought + Prolog code, respectively. The results reveal that the Prolog generation model surpasses the baseline in performance, while the combination generation models do not yield significant improvements. The Prolog corpus\footnote{\url{https://huggingface.co/datasets/Thomas-X-Yang/gsm8k-prolog}} based on GSM8K\footnote{\url{https://huggingface.co/datasets/gsm8k}} and the correspondingly finetuned Prolog generation model\footnote{\url{https://huggingface.co/Thomas-X-Yang/Llama-7b-gsm-prolog}} based on LLaMA7B\footnote{\url{https://huggingface.co/decapoda-research/llama-7b-hf}} are released to the research community.

\section{Introduction}
Presently, there exists a notable surge in interest in Natural Language Processing (NLP) catalyzed by the advent of ChatGPT. ChatGPT, being a transformer-based generative language model, exhibits versatility in performing a wide range of tasks grounded in natural language. The remarkable achievement of the GPT model can be attributed, in significant part, to its utilization of an exceptionally extensive corpus and a vast parameter set for acquiring features from the corpus. Nevertheless, mere augmentation in the model's size falls short in addressing mathematical inquiries encompassing arithmetic, common-sense, and symbolic reasoning – topics that may appear deceptively simple to individuals \cite{rae2022scaling}. One conceivable explanation for this issue is that generative models overly depend on their training corpus. Its proficiency in specific tasks stems from the presence of sentences closely, and sometimes explicitly, linked to those tasks within the corpus. The GPT model, in turn, assimilates these sentences and effectively memorizes the corresponding answers. However, mathematical problems pose a challenge as they can manifest in various contextual frames, articulated through diverse approaches, and involve distinct numerical values. Consequently, the model is prone to encountering ostensibly unfamiliar mathematical queries, resulting in suboptimal performance.

Therefore, the significance of this project hinges upon the inadequacy of current capabilities in addressing this type of questions. We posit that to surmount this limitation, an NLP model should possess the capability to ingest natural language sentences and produce corresponding logical predicates. These predicates can then be processed by an external tool, distinct from a language model, to ultimately compute the desired result. In this context, we employ the Prolog language, known for its efficacy in performing such tasks, as the external logic tool. In essence, the role of the language model is restricted to semantic parsing and question comprehension, while the logical and computational tasks are delegated to a more precise tool. In this manner, the language model is relieved of the burden of memorizing every conceivable answer to questions, focusing instead on proficiently translating natural language into logic language. This shift in approach has the potential to significantly diminish the model's reliance on an excessively large corpus and enhance its performance in tackling such questions.

Furthermore, a significant issue with neural networks is the limited space for human intervention. It poses a challenge for humans to comprehend the inner workings and exert control over a vast neural network. Through the adoption of this model paradigm, human involvement is facilitated through the control of an external tool responsible for executing logic language, thus augmenting the model's explainability.

The paper's structure is as follows: The concept of chain-of-thought will be discussed in the Related Work section. Subsequently, the Approach section will provide a detailed, step-by-step account of project implementation. The Results section will showcase the performance of the fine-tuned models. Lastly, the Conclusion section presents drawn conclusions and outlines future research directions.

\section{Related Work}
In prior research, chain-of-thought prompting has demonstrated its efficacy in enhancing the reasoning capabilities of large language models when compared to conventional prompts that only supply questions and answers \cite{wei2023chainofthought}. The fundamental concept behind chain-of-thought is to elucidate the intermediate problem-solving steps to the model. Consequently, we are motivated to employ chain-of-thought in finetuning as a foundational benchmark for this project, with the aim of investigating whether fine-tuning for Prolog generation as the output surpasses fine-tuning for chain-of-thought as the output in terms of performance.

\section{Approach}
Initially, this study leverages ChatGPT in conjunction with human correction to acquire Prolog code for each sample within the GSM8K dataset. It then employs the Chinese-Vicuna framework to fine-tune the LLaMA model across four data output configurations. Ultimately, the study assesses performance under these configurations to gauge the effectiveness of Prolog code generation for solving mathematical problems.

\subsection{Base Corpus}
The selection of GSM8K as the foundational corpus is based on its high relevance and quality. GSM8K comprises more than 8.5k elementary school-level math problems along with their solutions, articulated in natural language \cite{cobbe2021training}. Each solution employs a chain-of-thought approach to address the question, presenting a step-by-step solution culminating in a final answer, which can already be directly used as one output style.

\subsection{Prolog Code Retrieval}
\begin{figure*}[htp]
    \centering
    \includegraphics[width=14cm]{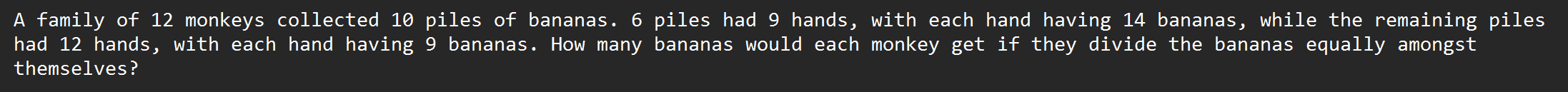}
    \caption{One example of a question in GSM8K}
    \label{fig:figure 1}
\end{figure*}

\begin{figure*}[htp]
    \centering
    \includegraphics[width=14cm]{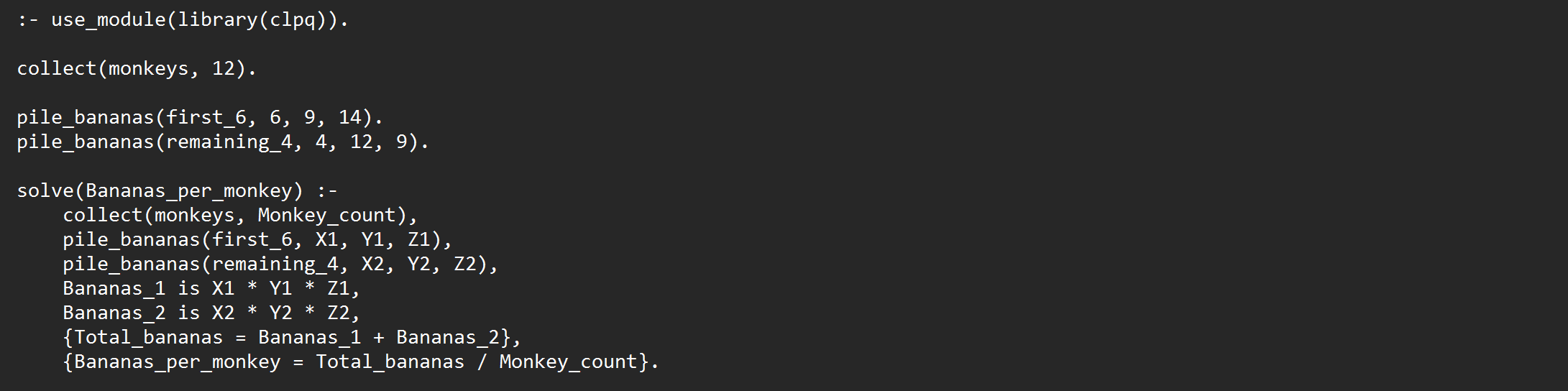}
    \caption{One example of a piece of Prolog code}
    \label{fig:figure 2}
\end{figure*}

We formulate prompts to extract Prolog code from ChatGPT. We first manually compose 10 examples for integration into the few-shot prompts. Here, we present an illustrative example. Figure \ref{fig:figure 1} depicts a question from GSM8K, while Figure \ref{fig:figure 2} showcases the corresponding Prolog code designed to address it.

In an effort to enhance the accuracy of retrieval results, the prompts incorporate natural language answers from GSM8K. Initially, the gpt-3.5-turbo-16k model is employed due to its cost-effectiveness. We process a pool of 100 samples, selecting 20 of them to construct the prompts, maximizing the utilization of the input token length. Utilizing the newly crafted prompts, we generate codes for all samples, retaining those codes that are both executable and yield accurate outcomes and regenerating codes for the rest. This iterative process continues until a bottleneck is encountered.

To overcome this bottleneck, we inject randomness into the process by reconfiguring the prompt candidates. The revised prompts comprise two components: the fixed part and the random part. The fixed part retains 8 old candidates. We select the 64 longest correctly generated code pieces, as longer codes often encompass more intricate arithmetic operations and potentially contribute to improved correctness. For each sample, we randomly select five candidates from this group to constitute the random part. This generation process continues iteratively until the bottleneck is encountered once again.

By this stage, the number of remaining questions has dwindled considerably, allowing us to employ gpt-4, which has the potential to further decrease the number of remaining questions to fewer than 100.

Finally, the remaining pieces of Prolog code are finalized through manual completion, followed by a manual verification of both executability and correctness of all the codes.

\subsection{Finetuning}
Owing to VRAM limitations, we employ LoRAs to facilitate the fine-tuning of the LLaMA7B model within the Chinese-Vicuna framework\footnote{\url{https://github.com/Facico/Chinese-Vicuna}}. While the inputs consist of mathematical problems, we explore four different output styles for fine-tuning: chain-of-thought, Prolog code, chain-of-thought + Prolog code, and Prolog code + chain-of-thought. This approach yields four fine-tuned 7B models. Throughout the fine-tuning process, the same configuration of hyperparameters is maintained to ensure a fair comparison of the performance of the four models.

\section{Results}
All four fine-tuned models undergo performance testing using the identical test set. Beam search is chosen as the generation strategy due to its superior performance in comparison to random sampling. Table \ref{tab:table 1} and Table \ref{tab:table 2} present the accuracy results for chain-of-thought and Prolog code, respectively. In the case of a chain-of-thought result, a syntax error is identified when the parser fails to retrieve an answer, whereas a semantic error occurs when the retrieved answer is incorrect. In the case of a Prolog code result, a syntax error indicates non-executability of the code, whereas a semantic error signifies that the executable code produces an incorrect answer.

\begin{table*}[htp]
\centering
\begin{tabular}{llll}
Finetuning Data       & Accuracy & Syntax Error & Semantic Error \\ \hline
GSM                   & 25.1\%   & 2.5\%        & 72.4\%         \\ \hline
GSM Prolog            &          &              &                \\ \hline
GSM Prolog (COT+Code) & 16.1\%   & 47.8\%       & 36.0\%         \\ \hline
GSM Prolog (Code+COT) & 26.2\%   & 3.2\%        & 70.7\%         \\ \hline
\end{tabular}
\caption{The chain-of-thought part performance of four finetuned models}
\label{tab:table 1}
\end{table*}

\begin{table*}[htp]
\centering
\begin{tabular}{llll}
Finetuning Data       & Accuracy & Syntax Error & Semantic Error \\ \hline
GSM                   &          &              &                \\ \hline
GSM Prolog            & 30.9\%   & 20.6\%       & 48.4\%         \\ \hline
GSM Prolog (COT+Code) & 17.7\%   & 59.1\%       & 23.2\%         \\ \hline
GSM Prolog (Code+COT) & 30.1\%   & 24.9\%       & 45.0\%         \\ \hline
\end{tabular}
\caption{The Prolog code part performance of four finetuned models}
\label{tab:table 2}
\end{table*}

\subsection{Finetuned on Chain-of-Thought}
This model results from the fine-tuning of LLaMA7B directly using the natural language answers from GSM8K. Earlier research revealed that LLaMA7B, prior to fine-tuning with the math corpus, achieved an accuracy of 11.0\% on GSM8K's test set \cite{touvron2023llama}. Following fine-tuning, LLaMA7B exhibits an accuracy of 25.1\%, surpassing its previous performance. The increment in performance can be attributed to both fine-tuning and the adoption of the beam search generation strategy. This suggests that introducing chain-of-thought samples during the fine-tuning phase can positively impact the model's performance. We consider this performance as the baseline.

\subsection{Finedtuned on Prolog Code Generation}
This model undergoes fine-tuning to produce Prolog codes for mathematical questions. The generated Prolog codes are subsequently forwarded to a Prolog compiler for correctness verification. Remarkably, this fine-tuned Prolog generation model attains an impressive accuracy of 30.9\%, a substantial improvement over the baseline. This suggests that entrusting the logical and computational aspects to an external tool and relegating the model's role to that of a translational device can effectively enhance its performance in solving math problems that necessitate logical and computational inference.

\begin{figure*}[htp]
    \centering
    \includegraphics[width=14cm]{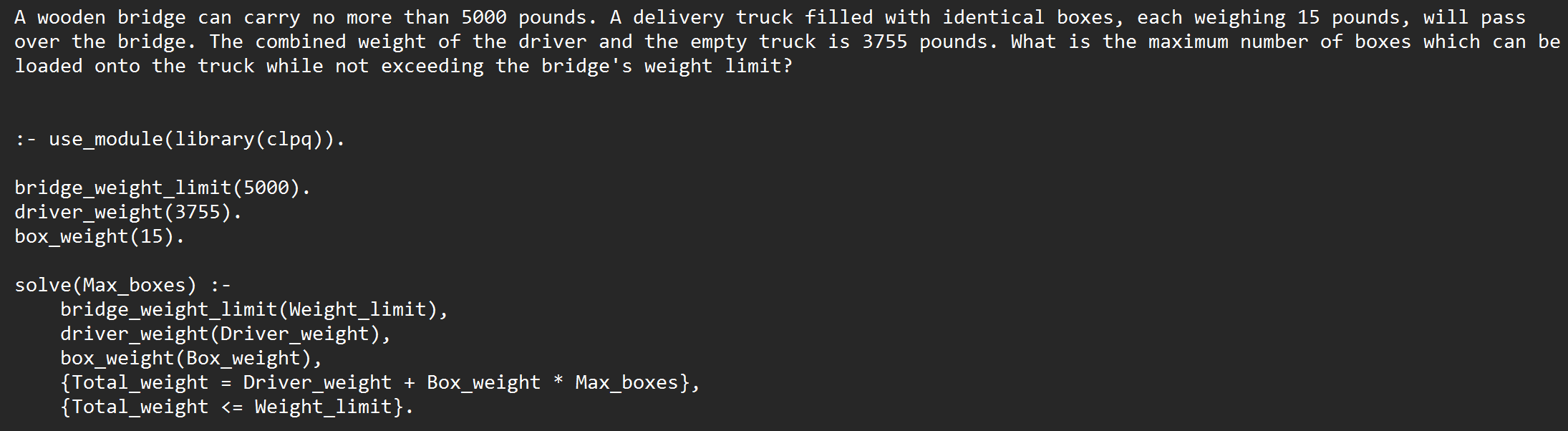}
    \caption{One example of a piece of Prolog code with syntax errors}
    \label{fig:figure 3}
\end{figure*}

Nonetheless, it has come to our attention that certain outputs categorized as having syntax errors do not necessarily entail critical errors involving ambiguity or semantic flaws. These outputs incorporate operations that the compiler does not support, rendering the codes non-executable. Such issues can be rectified by expanding the compiler's capabilities to encompass a broader range of operations. Therefore, it may be overly stringent to classify these outputs as incorrect. An illustrative example is provided in Figure \ref{fig:figure 3}, where the error arises due to the problem of integer solutions to an inequality, which the compiler cannot handle accurately.

Following manual review, approximately 1.5\% of the samples are reclassified as correct when the criteria are relaxed, considering the type of samples mentioned earlier as correct. As illustrated in Table \ref{tab:table 3}, the accuracy now stands at 32.4\%. While the improvement may not be substantial, it underscores the significance of the reliability of the external tool.

\begin{table*}[htp]
\centering
\begin{tabular}{llll}
Finetuning Data      & Accuracy & Syntax Error & Semantic Error \\ \hline
GSM Prolog           & 30.9\%   & 20.6\%       & 48.4\%         \\ \hline
GSM Prolog (Revised) & 32.4\%   & 19.1\%       & 48.4\%         \\ \hline
\end{tabular}
\caption{The revised Prolog code part performance of the Prolog model}
\label{tab:table 3}
\end{table*}

\subsection{Finetuned on Chain-of-Thought + Prolog Code}
This finetuned model generates a chain-of-thought solution followed by a piece of Prolog code. This experiment is motivated by the fact that transformer models utilize the current sequence to generate subsequent tokens, implying that the content generated initially can influence subsequent content generation. Our objective is to assess whether generating chain-of-thought solutions first can improve the accuracy of the generated codes. The experiment's outcome reveals that this combination not only diminishes the performance of the chain-of-thought but also lowers the code accuracy to as low as 17.7\%, a level even below the baseline. One contributing factor is that this output combination contaminates the data, making it challenging for the model to discern the relationships between tokens during the fine-tuning phase. Another potential explanation is that chain-of-thought may not inherently enhance the quality of code generation during the inference stage.

\subsection{Finetuned on Prolog Code + Chain-of-Thought}
The motivation of this experiment aligns with that of the combination experiment in the previous section. Our objective is to investigate the impact of Prolog codes on chain-of-thought generation. Interestingly, when Prolog code generation is not influenced by chain-of-thought this time, its quality, achieving an accuracy of 30.1\%, closely approximates that of solely generating Prolog codes. Furthermore, code generation appears to exert a marginal, positive influence on chain-of-thought generation that follows. As a result, the accuracy of chain-of-thought rises to 26.2\%. This observation suggests that generations characterized by a clear and easily comprehensible structure may aid the model in extracting information that benefits subsequent generated content.

\section{Conclusion}
In this study, we utilized ChatGPT to generate Prolog codes for the GSM8K corpus, employing four distinct output settings and subsequently fine-tuning four LLaMA7B models accordingly. Through a comparative analysis of accuracies on the test set, the following conclusions can be drawn. Firstly, fine-tuning undeniably enhances performance in the mathematics question domain. Secondly, the process of generating Prolog codes and subsequently sending them to an external compiler yields superior results compared to chain-of-thought generation. Leaving the logical and computational aspects to an external tool and reducing the model to a translational device can effectively enhance its performance in solving math problems. This approach has the potential to be applied to other domains that involve logical and computational reasoning. Thirdly, generating combinations of chain-of-thought and Prolog code does not result in a significant further enhancement of inference performance.

So far, the finetuned LLaMA7B for Prolog generation continues to exhibit a 19.1\% syntax error rate after revision and a 20.6\% before revision. Therefore, future efforts can be directed towards reducing syntax errors and expanding the capabilities of the external tool to support additional operations. Furthermore, given the remarkably high semantic error rate of 48.4\%, there is a pressing need for models with enhanced question comprehension capabilities to address this bottleneck.

\section{Acknowledgement}
Special thanks are given to Professor Yik-Cheung Tam for mentoring this project, and to NYU Shanghai for providing the platform to support undergraduate research.

\newpage
\bibliographystyle{IEEEtran}
\bibliography{references}

\end{document}